\documentclass[10pt,twocolumn,letterpaper]{article}

\usepackage{wacv}
\usepackage{times}
\usepackage{epsfig}
\usepackage{graphicx}
\usepackage{amsmath}
\usepackage{amssymb}
\usepackage{multirow}
\usepackage{pifont}
\usepackage[accsupp]{axessibility}  

\def\wacvPaperID{****} 

\wacvfinalcopy 

\ifwacvfinal
\fi

\ifwacvfinal
\usepackage[breaklinks=true,bookmarks=false]{hyperref}
\else
\usepackage[pagebackref=true,breaklinks=true,colorlinks,bookmarks=false]{hyperref}
\fi

\begin{document}

\title{Towards Active Vision for Action Localization with Reactive Control and Predictive Learning}

\author{Shubham Trehan\\
Department of Computer Science\\
Oklahoma State University\\
{\tt\small strehan@okstate.edu}
\and
Sathyanarayanan N. Aakur\\
Department of Computer Science\\
Oklahoma State University\\
{\tt\small saakurn@okstate.edu}
}

\maketitle

\ifwacvfinal
\thispagestyle{empty}
\fi

\begin{abstract}
Visual event perception tasks such as action localization have primarily focused on \textit{supervised} learning settings under a \textit{static} observer, i.e., the camera is static and cannot be controlled by an algorithm. They are often restricted by the quality, quantity, and diversity of \textit{annotated} training data and do not often generalize to out-of-domain samples. In this work, we tackle the problem of \textit{active} action localization where the goal is to localize an action while controlling the geometric and physical parameters of an active camera to keep the action in the field of view \textit{without training data}. We formulate an energy-based mechanism that combines predictive learning and reactive control to perform active action localization \textit{without rewards}, which can be sparse or non-existent in real-world environments. We perform extensive experiments in both simulated and real-world environments on two tasks - active object tracking and active action localization. We demonstrate that the proposed approach can generalize to different tasks and environments in a streaming fashion, without explicit rewards or training. We show that the proposed approach outperforms unsupervised baselines and obtains competitive performance compared to those trained with reinforcement learning. 
\end{abstract}
\vspace{-1.5em}
\section{Introduction}
\begin{figure}[t]
    \centering
    \includegraphics[width=0.99\columnwidth]{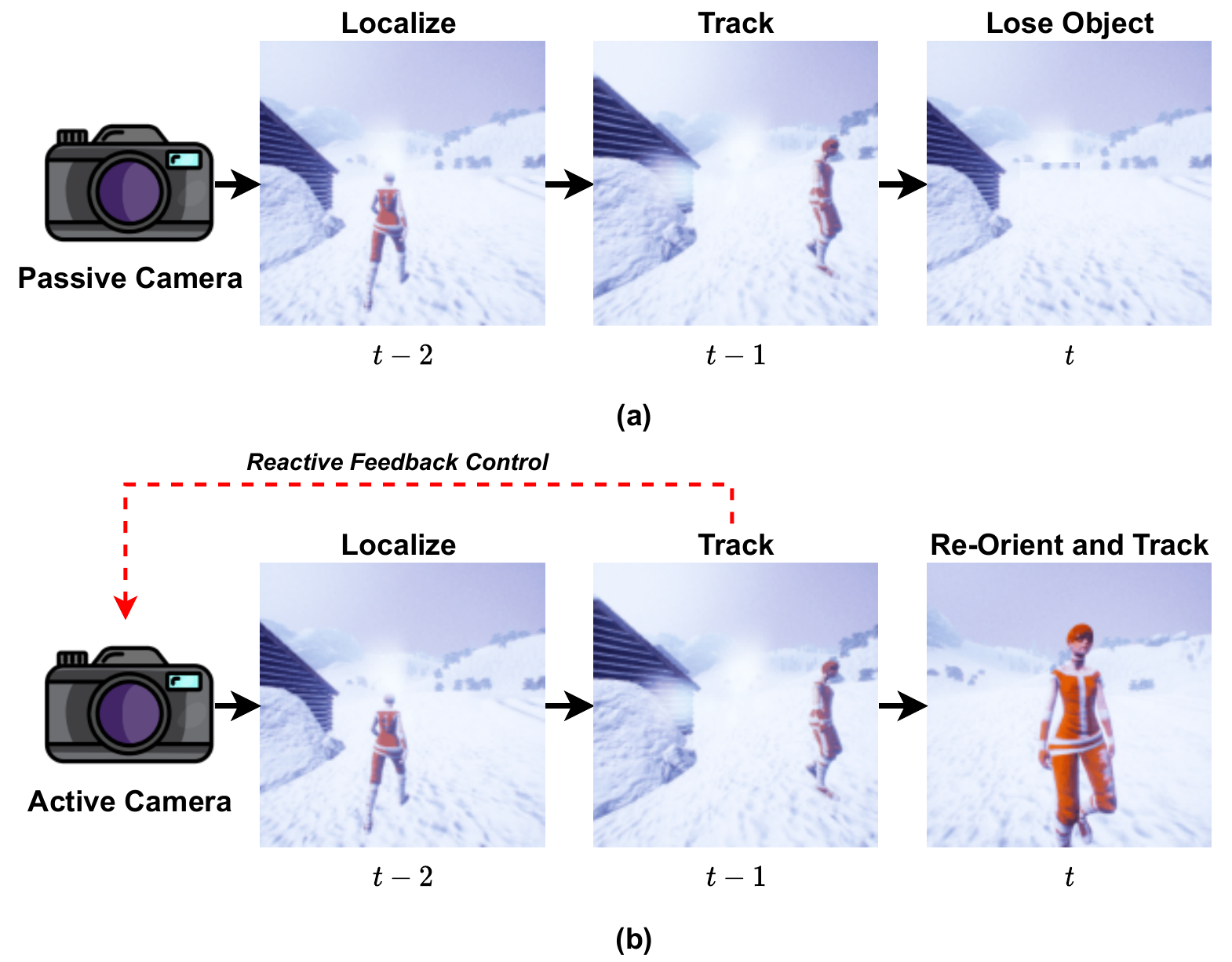}
    \caption{Illustration of a passive vision framework (a) vs an \textbf{active vision} framework (b) for action localization. Without continual, reactive feedback control, a passive camera can lose track of a moving actor due to motion, occlusion or noise.}
    \label{fig:intuition}
\end{figure}

Advances in visual event understanding tasks such as action recognition, segmentation, and localization have largely been driven by the assumption that the observer is \textit{passive} i.e., the algorithm or agent cannot manipulate the geometric and physical parameters of the sensors such as fixation and self-motion. This severely restricts the ability of the observer to view, understand and respond to the events in the scene to improve the quality of the eventual perceptual results such as recognition or localization. For example, consider the scenario in Figure~\ref{fig:intuition}, where an actor (or object) of interest is being observed and moves out of the field of view of the observer. A passive agent cannot re-position itself to continue to observe the action to understand the event and must wait for the actor to return within its field of view. An \textit{active} vision system, on the other hand, can re-orient itself by controlling the geometric and physical parameters of the camera and continue to observe the event. Similarly, an active vision system can overcome such challenges when faced with occlusions to perform its underlying task without constraints. Note that active vision is different from \textit{active sensing}, where the sensor probes its environment with self-generated energy. 

Inspired by cognitive theories of perception~\cite{horstmann2015surprise,horstmann2016novelty, zacks2001event}, we formulate an energy-based framework based on the idea of surprise modeling and reactive control. Spatial-temporal surprise has shown to be an underlying factor in both attention~\cite{aakur2020unsupervised,horstmann2015surprise} and event perception~\cite{aakur2019perceptual, zacks2001event} while reactive control~\cite{brooks1986robust,crowley1994integration} has been argued to be essential in real-time control of mobile, active vision systems. 
We leverage these characteristics to formulate and implement an unsupervised, energy-based framework to localize \textit{generic} actions (i.e., not constrained to a domain, actor-type, or action semantics) in an environment using spatial-temporal surprise and actively control the geometric parameters of the camera using reactive control \textit{without any rewards or training needs}. 
Our perception framework continuously observes, predicts, and learns spatial-temporal attention maps to locate the action of interest, while the reactive control model uses this to re-orient itself to keep the action in its field of view to overcome occlusion, fast movement, and observational noise. Our framework works in a \textit{reactive, streaming} manner and works in real-time to localize, track and build robust representations of the scene. 

In this work, we tackle the problem of \textit{active} action localization through self-supervision in real-world scenarios. Specifically, we look at action localization, where the goal is to identify and localize an action of interest in a given scene. Traditional approaches to action localization \cite{aakur2021learning, duarte2018videocapsulenet, wang2018action} have focused mainly on spatial-temporal localization in static camera settings, where the goal is to localize actions within the camera's fixed field of view (FOV). We consider the scenario where the algorithm has access to the camera's geometric and physical parameters and can manipulate the camera to focus on the action of interest as it moves within and out of its field of view. 
While tremendous progress has been made in embodied visual agents \cite{DBLP:journals/corr/abs-1711-11543, sdk2021}, it has primarily been driven by simulation environments \cite{kolve2019ai2thor, habitat19iccv, shenigibson} where the reward is easily measured and the agent is the major active entity. In real-world tasks, such rewards can be very sparse or even non-existent and hence requires a flexible, reactive framework to adapt to function in environments with evolving scenarios and actors. 

\textbf{Contributions:} The contributions of our approach are three-fold: (i) we are among the first to tackle \textit{active} action localization in streaming videos without the need for extensive training or simulation environments with hardwired rewards functions, (ii) we show that predictive learning can be combined with reactive control systems to perform self-supervised active action localization, and (iii) we show that the approach can generalize to both synthetic and real-world environments without \textit{training} and demonstrate its efficiency on two challenging tasks in active object tracking and active generic action localization.

\section{Related Work}
\textbf{Active vision} systems for action understanding have been an understudied problem in computer vision literature, primarily due to the complexity of active control and nature of actions. Common approaches to active vision systems for actions~\cite{alzarok20173d,ccelik2017color,dickinson1997active} impose conditions on the color and shapes of objects that can be actively tracked. Daniilidis \textit{et al.}~\cite{daniilidis1998real} impose conditions on the motion of the scene by restricting the environment to the motion of a single object in the field of view as a cue with a stationary background. 
Wilkes \textit{et al.}~\cite{wilkes1993behaviors} use a single camera on the robotic arm to move to a viewpoint chosen based on three non-collinear feature points of an object. Dickinson \textit{et al.}\cite{dickinson1993integrating} reconstructs the surrounding environment of a robot by using a recognition pipeline and a "Planner" pipeline. The recognition pipeline recognizes the object around it, and the planner moves the camera head in the direction for a better view.

Control-based approaches~\cite{ri2017image,spurlock2015discriminative} use image-based features to extract object pose and relative camera poses to orient the camera towards the actor of interest. Advances in realistic simulation environments~\cite{kolve2019ai2thor, habitat19iccv, shenigibson} have enabled the use of reinforcement learning agents~\cite{huang2019learning, kyrkou2020imitation, li2020pose, DBLP:journals/corr/LuoSML17,wang2020active} for building active vision systems for more complex actions, objects and environments. Wenhan \textit{et al.}~\cite{DBLP:journals/corr/LuoSML17} achieved success in active tracking using a CNN-LSTM architecture. Huang \textit{et al.}~\cite{huang2019learning} use videos containing only one subject and imitates the actions of a professional cameraman in a \textit{supervised} setting. Kyrkou \textit{et al.} \cite{kyrkou2020imitation} proposes to estimate the motion displacement of the camera and the number of targets in the image using regression tasks. Wang \textit{et al.}~\cite{wang2020active} use RNNs to account for unobserved frames for camera selection. Jing \textit{et al.}~\cite{li2020pose} use multiple cameras for active tracking, with vision-based and pose-based controllers.

\textbf{Action localization} has been a widely studied area of research, primarily tackled through supervised learning~\cite{gkioxari2015finding,hou2017tube,jain2017tubelets,soomro2015action,soomro2016predicting}. For localization in streaming videos, predictive learning\cite{aakur2020action,aakur2021learning} has been used in self-supervised action localization by constructing attention-based feature representation. 
Xie \textit{et al.}~\cite{xie2018correlation} propose the use of multiple correlations filter (CF) models, controlled using a multiplexer trained with reinforcement learning.
Wang \textit{et al.}\cite{wang2018action} leverage 3D skeleton sequences by tracking the human joints to recognize actions. Duarte \textit{et al.}~\cite{duarte2018videocapsulenet} use a 3D capsule network in videos to do pixel-wise action segmentation.

\section{Active Generic Action Localization}

\begin{figure*}
    \centering
    \includegraphics[width=0.85\textwidth]{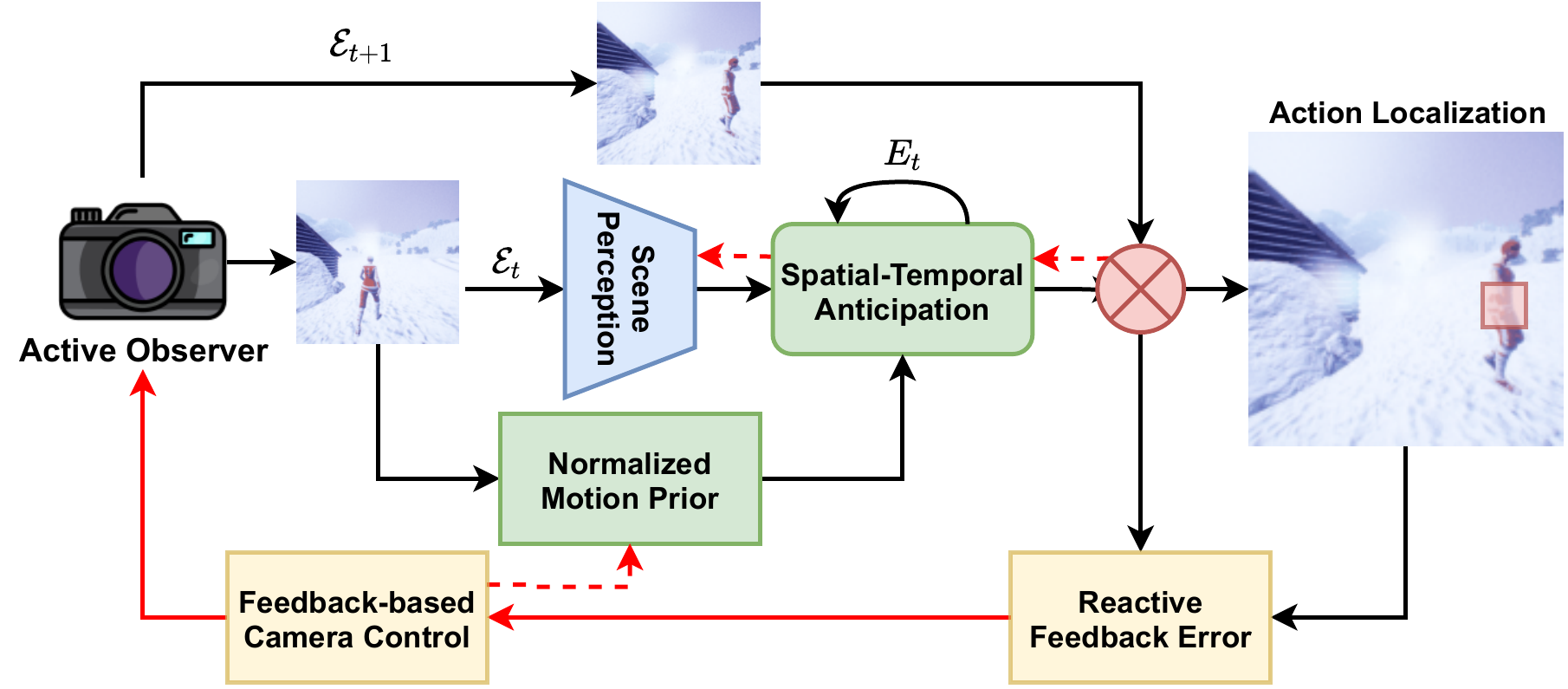}
    \caption{\textbf{Overall Architecture}. There are three major components - scene perception (blue), modeling temporal dynamics with spatial-temporal anticipation (green) and reactive feedback control (yellow) for active perception. The spatial-temporal error is used for real-time camera control and action localization. Black line - forward pass, red dashed lines - learning signal and red solid lines - control feedback.}
    \label{fig:arch}
\end{figure*}

\textbf{Problem Setup.} Consider a scene or environment $\mathcal{E}_t$ observed at time $t$ by an active camera $c$. The camera observes the scene in a \textit{streaming} fashion, i.e., the frames are received sequentially, and as such, it has no access to any future states or observations of the scene. The goal is to observe the scene localize the \textit{dominant} action $a_i\in a_n$, as observed by the camera $c$, and manipulate the camera position to ensure that the actor is within the camera's field of view. 
The key challenge is to ignore clutter and identify the object of interest (i.e., the \textit{actor}) \textit{without any supervision} while modeling the relational dynamics of the scene to enable active camera control. We aim to construct a unified, self-supervised framework based on the idea of \textit{predictive learning} for self-supervised active action localization.

\subsection{Overview} 
We formulate our framework around the idea of error minimization using predictive learning. The overall approach is illustrated in Figure~\ref{fig:arch}. We aim to minimize two different errors - the predictive error from modeling the spatial-temporal dynamics and a reactive error due to camera orientation. The key idea is to \textit{locate} the action of interest, i.e., the \textit{actor} through perceptual predictive learning and maintains the object within the field of view by minimizing the error between the location of the action and the camera center through reactive error. The overall learning objective then becomes the minimization of the error-driven energy function given by
\begin{equation}
    E(y) = -\ln p(\mathcal{E}_t\lvert\theta, E_t, c) + \lvert\lvert q(a) - c(a) \lvert\lvert^2
    \label{eqn:freeEnergy}
\end{equation}
where $\mathcal{E}_t$ is the observed scene, $E_t$ is an internal representation of the observed action ($a$), $c$ is the active camera orientation and $q(a)$ is the actual location of the perceived action and $c(a)$ is the observed location of the action as observed by the active camera; $\theta$ represents the learnable parameters of the proposed framework. 
Since computing the likelihood of observing the current scene can be intractable, we model the first term as the minimization of \textit{surprise} encountered by the underlying sensory (perception) modules (Section~\ref{sec:percep_pred}). Higher prediction errors indicate lower probability of observing the scene and hence minimizing the prediction errors allow us to capture the  probability of observing the scene in a more tractable manner. The second term is the divergence between the actual position of the action and the state inferred by the internal states of the system (Section~\ref{sec:pid}). A stable state is reached when this energy is minimal, i.e., the inferred actor location approximates the actual, observed location (second term), and the perceptual surprise is minimal (first term). The system adapts to the changes in the observed data to maintain a stable state by minimizing the overall error-driven energy.

\subsection{Perceptual Prediction}\label{sec:percep_pred}
The first step in our framework is to identify an action of interest and localize it spatially. Inspired by cognitive theories of event perception \cite{horstmann2015surprise, zacks2001event}, we follow prior work \cite{aakur2019perceptual, aakur2020action, aakur2021learning}, and use the idea of predictive learning for generic action localization. 
We begin with global, scene-level feature extraction using a convolutional neural network. Note that unlike \cite{aakur2020action, aakur2021learning} we do not use object detection due to the overhead induced. Instead, we use the output of the fourth convolutional block ($f^S_t$) as the spatial feature representation of the observed scene at time $t$. Second, we model the spatial-temporal dynamics of the scene using a stack of LSTM \cite{hochreiter1997long} networks and construct an internal representation $E_t$ of the observed event. We use an attention mechanism \cite{bahdanau2016neural} $\alpha^S_t$ to weight spatially relevant features that can help in anticipating areas of interest in subsequent time steps. 
Formally, we define the perceptual prediction model as a prediction function that maximizes the probability of observing the future spatial feature $\hat{f}^S_{t+1}$ conditioned on the current observed features $f^S_t$, an internal representation $E_t$, an attention function $\alpha_t$ and a set of learnable weights $\theta$. We minimize the prediction error given by 
\begin{equation}
    \mathcal{L}_{global} = {\lVert f^S_{t+1} - {f}^S_{t}\lVert_{2}} \odot {\sigma(f^S_{t+1} - {f}^S_{t})} 
    \label{eqn:globalPred}
\end{equation}where $\sigma(\cdot)$ is the sigmoid activation function to scale the motion between $0$ and $1$. This forces the network to suppress spurious motion patterns that can arise due to slight camera motion and background noise. Note that this is different from both \cite{aakur2020action} (which uses a first order hold) and \cite{aakur2021learning} (which uses the L2-norm between the predicted and observed features at time $t+1$) to weight the prediction error. 
We use the prediction errors and construct a spatial-temporal error map given by $\hat{\alpha}_t = \frac{\exp(e_{ij})}{\sum_{m=1}^{w_k}\sum_{n=1}^{h_k}\exp(e_{mn})}$, where $e_{ij}$ refers to the prediction error at location $(i,j)$ of $\mathcal{L}_{global}$. We use $\hat{\alpha}_t$ for localizing the actions of interest since it accounts for the quality of predictions (i.e., surprise minimization) and the relative spatial alignment of the prediction errors. The location with the maximum prediction error is considered to be the location of the dominant action.

\textbf{Prediction-guided Attention Feedback.} In addition to the prediction loss objective defined in Equation~\ref{eqn:globalPred}, we also allow for a feedback error from the action localization cue $\hat{\alpha}_t$ to guide the spatial attention ($\alpha_{t_1}$) at  time $t+1$. We force the attention mechanism to align with the previous error-based attention by minimizing the error function given by 
\begin{equation}
\mathcal{L}_{attn} = {\lVert \alpha_{t} - \hat{\alpha}_{t-1}\lVert_{2}} + {\lVert \alpha_{t}\odot{f^S_t} - \hat{\alpha}_{t-1}\odot{f^S_{t}}\lVert_{2}}
    \label{eqn:attentionError}
\end{equation}
where the first term forces the attention at time $t$ to match the spatial-temporal prediction error map from the previous time step and the second term forces the attention-weighted features from $t-1$ to match the observed features at $t$. We find that this additional loss function enforces temporal consistency in spatial-error prediction and hence allows for a more coherent localization, especially under an active vision setup. Hence the overall training objective is a combination of global prediction error $\mathcal{L}_{global}$ and the attention-based feedback loss $\mathcal{L}_{attn}$. Empirically, in Section~\ref{sec:ablation}, we find that these two losses improve the performance for both active object tracking and active action localization tasks. 

Note that our approach is not explicitly trained for active localization using rewards or annotations, the perceptual prediction model continuously predicts, compares and attends to the action of interest in the scene. To help adapt to the changing scene dynamics, the CNN encoder and the prediction stack are constantly updated at every time instant $t$. This constant update allows the model to adapt to changes in the motion of the agent, the object of interest and the environment to provide cues for the active camera control.

\subsection{Active Control with Reactive Modeling}\label{sec:pid}
The final step in our active vision framework is the continual control of the embodied agent equipped with the camera. We use a reactive control formulation to adjust the camera orientation by minimizing the distance ($e_{ctrl}(t)$) between the action location $q_t(a)$ and the current camera position $c_t(a)$ at any time instance. We use the term \textit{reactive} control since the behavior of the active camera is based on \textit{events} formulated based on the spatial-temporal prediction error map $\hat{\alpha}_t$ and not discriminative or anticipatory. Formally, we represent the camera control mechanism as an energy minimization term given by
\begin{equation}
    c_{t+1} = \lambda_p e_{ctrl}(t) + \lambda_d \frac{d}{dt}(e_{ctrl})
    \label{eqn:pid}
\end{equation}
where $e_{ctrl}$ is the distance between the ideal camera position and the current position. The first term refers to the action taken by the active camera to account for prediction errors and re-align itself to track the action of interest. In contrast, the second term is used as ``anticipatory control'' to reduce the effect of the position error by choosing an action influence generated by the rate of change of the error. This energy term is inspired by the idea of PID controllers in control systems~\cite{araki2009pid}. Since we do not explicitly train for control, we define these distances as a relative angular distance such that the ideal position $q_t(a)$ is the action location derived from $\hat{\alpha}_t$ and the current position ($c_t(a)$) is the center of the camera's field of view. This formulation allows us to negate the need for training data and tightly couple perception and action spaces. The reactive control system continuously considers the prediction-based error map and the current camera position to ensure that the actor of interest is always within the camera's field of view, ideally at the center. In our experiments, we keep $\lambda_p=1$ and $\lambda_d=0.1$ and use the same formulation for all baselines.

\section{Experimental Evaluation}
\textbf{Implementation Details.} We use VGG-16~\cite{simonyan2015deep} (VGG-16) pre-trained on ImageNet \cite{5206848} as our CNN encoder to extract scene-level representations. The input to the encoder is of size $224\times 224$. The scene representation $f^S_t$ and hence the prediction error map $\alpha_t$ have a spatial dimension of $14\times 14$. We use three LSTMs in the prediction stack, with each successive LSTM using the output of the LSTM at the lower level. The number of hidden units in the LSTM prediction stack is set to 512. The learning rate was set to $1\times 10^{-6}$ and adaptive learning~\cite{aakur2019perceptual} was used to prevent overfitting. The evaluation was done on a server with an 32-core AMD Epyc CPU and an NVIDIA Titan RTX. 

\textbf{Evaluation Overview.} To evaluate our framework, we perform experiments on two tasks. First, we evaluate on the active object tracking benchmark, where the goal is to control an embodied agent with a static camera and track a moving object. In the second task, we construct a real-world benchmark evaluated on a physical robot arm equipped with a camera. The robot will be presented with a scene, and the goal is to track and localize the dominant action in the scene. Combined, the two experiments provide a comprehensive evaluation of the proposed framework to perform \textit{active} action localization under varying levels of difficulty across simulated and real-world environments.

\subsection{Active Tracking}\label{sec:aot}
\begin{table*}[ht]
    \centering
    \resizebox{0.99\textwidth}{!}{
    \begin{tabular}{|c||c|c||c|c|c|c|c|c|c|c|}
    \hline
        
         \multirow{2}{*}{\textbf{Approach}} & \multirow{2}{*}{\textbf{Generic}} & \multirow{2}{*}{\textbf{Training}}& \multicolumn{2}{|c|}{\textbf{\textit{SnowVillage}}} & \multicolumn{2}{|c|}{\textbf{\textit{DuelingRoom}}} & \multicolumn{2}{|c|}{\textbf{\textit{UrbanCity}}} & \multicolumn{2}{|c|}{\textbf{\textit{ParkingLot}}}\\\cline{4-11}
         & & & \textbf{Recall} & \textbf{Precision} & \textbf{Recall} & \textbf{Precision} & \textbf{Recall} & \textbf{Precision} & \textbf{Recall} & \textbf{Precision}\\
    \hline
    AD-VAT+\cite{zhong2018ad} & \ding{55} & \ding{51} (150K) & 0.71 & 0.51 & 0.95	& 0.69	& 0.97	& 0.71 & 0.67 & 0.46 \\
    \hline
    SmartTarget-RL\cite{zhong2018ad} & \ding{55} & \ding{51} (150K) & 0.63 & 0.42 & 0.85 & 0.62 & 0.94 & 0.70 & 0.62 & 0.38 \\
    \hline
    RandomTarget-RL\cite{zhong2018ad} & \ding{55} & \ding{51} (150K) & 0.60 & 0.41 & 0.65 & 0.55 & 0.97 & 0.69 & 0.41 & 0.58\\
    \hline
    \hline
    Active TLD & \ding{51} & \ding{55} & 0.11 & 0.01 (\textit{0.03}) & 0.15 & 0.02 (\textit{0.04}) & 0.18 & 0.01 (\textit{0.03}) & 0.10 & 0.00 (\textit{0.01}) \\\hline
    Active MIL & \ding{51} & \ding{55} & 0.12 & 0.01 (\textit{0.04}) & 0.13 & 0.01 (\textit{0.02}) & 0.15 & 0.02 (\textit{0.03}) & 0.11 & 0.01 (\textit{0.04}) \\\hline
    Active MOSSE & \ding{51} & \ding{55} & 0.09 & 0.01 (\textit{0.03}) & 0.14 & 0.01 (\textit{0.03}) & 0.17 & 0.02 (\textit{0.04}) & 0.13 & 0.02 (\textit{0.03}) \\\hline
    \textbf{Ours} & \ding{51} & \ding{55} & 0.68 & 0.16 (\textit{0.36}) & 0.85 & 0.26 (\textit{0.41}) & 0.75 & 0.14 (\textit{0.40}) & 0.49 & 0.23 (\textit{0.46}) \\
    \hline
    \end{tabular}
    }
    \caption{\textbf{Active Tracking Results}. We compare against reinforcement learning baselines trained specifically for tracking (indicated in the generic column). We outperform other generic baselines and perform competitively with supervised models on both metrics. Metrics for unsupervised models without distance penalty are in parentheses. Cumulative metrics per episode averaged over 100 runs are reported.}
    \label{tab:active_tracking_comp}
\end{table*}
First, we evaluate the framework on the active object tracking task. We follow prior work~\cite{zhong2018ad} and use a high-fidelity simulator to build 3D environments to simulate real-world active tracking scenarios. The environments simulate a photo-realistic world using the Unreal Engine, which is interfaced with UnrealCV~\cite{qiu2017unrealcv} and OpenAI Gym~\cite{brockman2016openai} using a wrapper function~\cite{zhong2017gym}. 
We use the same simulation environments as AD-VAT+~\cite{zhong2018ad} for fair comparison.

\textbf{Metrics.} 
The tracking quality metric $ \gamma_t $ is used to measure the relative position error based on a polar coordinate system, where the tracker is at the origin $(0, 0)$. We follow prior work~\cite{zhong2018ad} and define the tracking quality $ \gamma_t $ as 
\begin{equation}
\gamma_t = 1 - \lambda\frac{\lvert\rho_1 - \rho_2 \lvert}{\rho_{max}} - \lambda\frac{\lvert\theta_1 - \theta_2 \lvert}{\theta_{max}}
    \label{eqn:tracking_metric}
\end{equation}
where $(\rho_1, \theta_1)$ and $(\rho_2, \theta_2)$ are the current position of the tracker and the ideal position of the tracker, respectively; $\rho$ is the distance to the tracker, $\theta$ is the relative angle to the front of the tracker. $\rho_{max}$ is the maximum observable distance ($250$, in our case) to the tracker, and $\theta_{max}$ is the maximum view angle (the Field of View (FoV)) of the camera model ($90^{\circ}$ in our case). We bound $ \gamma_t $ to be between $[-1,1]$ to avoid over-punishment when the object is far away from the expected position. This formulation allows us to evaluate the quality of tracking since the metric in Equation~\ref{eqn:tracking_metric} is equivalent to \textit{precision} in conventional tracking since it penalizes both location and orientation to account for navigation. Smaller numbers indicate that the tracker is performing rather poorly, and the tracking ``episode'' is terminated when the tracker loses the target (i.e., $\gamma_t=-1$) for an extended period of time. Hence, the duration of tracking (i.e., the \textit{episode length}) is akin to the \textit{recall} metric used in conventional tracking literature. We terminate an episode when the maximum number of tracking steps ($500$) is reached or if the tracker loses the target for ten consecutive timesteps. 

\textbf{Baselines.} We evaluate our framework on three different environments (\textit{SnowVillage}, \textit{DuelingRoom}, \textit{UrbanCity}) and compare it to three fully supervised, reinforcement learning-based baselines: (i) \textit{AD-VAT+}~\cite{zhong2018ad}, an RL-based tracker trained together with a learned target in an adversarial framework, (ii) \textit{RandomTarget-RL}, an RL-based tracker trained with a target whose actions are randomly sampled and (iii) \textit{SmartTarget-RL}, an RL-based tracker trained with a navigator-like target, which plans the shortest path to a pre-defined goal. To randomize the trajectories, the goal coordinate and the initial coordinate are randomly sampled. Note that all three baselines are specialized models trained specifically for active tracking, while ours is a generic, event-based reactive framework. We also establish an unsupervised baseline based on a simulated active tracker based on traditional, long-term trackers such as TLD~\cite{kalal2011tracking}, MOSSE~\cite{bolme2010visual} and MIL ~\cite{babenko2009visual}. We take the tracker's output and use a PID controller to move the embodied agent to keep the tracked object at the center of the field of view. 

\textbf{Quantitative Results.} 
We summarize our results in Table~\ref{tab:active_tracking_comp}. As can be seen, we perform competitively with more sophisticated, reinforcement learning-based active object tracking frameworks that are explicitly designed and trained for this task and domain. 
We find that our approach consistently has comparable recall to RL-based models across all environments, including the highly challenging \textit{ParkingLot} and \textit{SnowVillage} environments. The precision is somewhat lower, which is attributed to the distance constraint ($\rho_{max}$ in the first term Equation~\ref{eqn:tracking_metric}). RL baselines use this information in the reward function to learn specific behaviors to follow the actor at a distance of $250$. On relaxing it to a binary option given by $\rho = \begin{cases}
    1, & \frac{\lvert\rho_1 - \rho_2 \lvert}{\rho_{max}} \leq 1 \\
    0, & \text{otherwise}
\end{cases}
$, we find that the precision increases. This result indicates that our approach can keep the object in the field of view for a longer amount of time. Additionally, it should be noted that the AD-VAT+ and SmartTarget-RL baselines take at least 20,000 iterations to match our performance, while the RandomTarget-RL takes around 80,000 iterations. Our approach, on the other hand, is not explicitly trained for this task and does not require rewards for learning. 

We also compare it to another generic, unsupervised baseline with a simulated \textit{active} TLD tracker. As can be seen, we significantly outperform generic, unsupervised trackers. Qualitatively, we find that the TLD tracker-based active agent fails to recover from sharp, rapid movements from the target. For example, when the target makes a sharp turn away from its current trajectory, the tracker fails to account for the change and does not adjust well. The tracker then quickly loses sight of the target and terminates. This effect is more pronounced under the \textit{SnowVillage} and \textit{ParkingLot} environments, where the background and the target can be heavily affected by occlusion. 

On the other hand, the predictive learning-based approach can quickly recover from any such lapses and regain focus to continue tracking. We attribute this to the actor-specific prediction (from Section~\ref{sec:percep_pred}), which forces the framework to predict the areas of high motion (mostly belonging to the action/actor of interest) in addition to the overall scene dynamics. This effect is highlighted in the relatively reduced performance of our approach on the \textit{UrbanCity} environment, where there are many distractor elements such as reflections that increase the \textit{surprise} encountered and hence moved the focus away from the target. 

\begin{figure}
    \centering
    \begin{tabular}{ccc}
    \hline
        \multicolumn{3}{c}{\textbf{Simulated Environments}}\\
        \hline
         \includegraphics[width=0.12\textwidth]{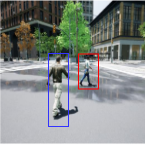} & \includegraphics[width=0.12\textwidth]{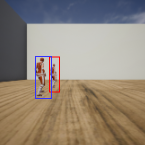} & \includegraphics[width=0.12\textwidth]{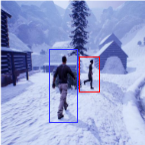} \\
         \textit{Urban City} & \textit{Dueling Room} & \textit{Snow Village} \\
         \hline
        \multicolumn{3}{c}{\textbf{Real World Setup}}\\
        \hline
        \includegraphics[width=0.12\textwidth]{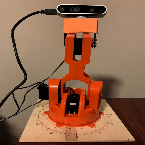} & \includegraphics[width=0.12\textwidth]{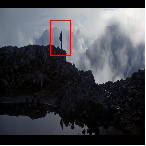} & \includegraphics[width=0.12\textwidth]{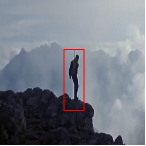} \\
        \textit{Active Camera} & \textit{Overall Scene} & \textit{Camera FOV}\\
    \end{tabular}
    \caption{Examples of the \textbf{evaluation environments} are illustrated here. The top row shows the simulated environments (target is highlighted in red and the embodied tracker is in blue). The bottom row shows the experimental setup for real-world environments. 
}
    \label{fig:environment}
\end{figure}
\subsection{Real-world Active Generic Action Localization}
In addition to evaluating on the active tracking benchmark, we also evaluate the ability of our framework to perceive a given scene, identify a salient action and track the action in \textit{real-world} scenarios. The key idea behind these experiments is to evaluate if the framework can generalize beyond simple, single-person, or single activity environments currently evaluated for active object tracking. 

\textbf{Evaluation Environment.} To test the framework beyond simulation, we construct a physical setup with an active camera for evaluating generic action localization. In place of a simulated, embodied agent, we formulate our framework to control an active camera setup embodied by an Arduino Braccio robotic arm~\cite{braccio} that is connected to an Intel RealSense Camera~\cite{grunnet2018projectors}.
The robot arm has three degrees of freedom and is mounted on a static surface, and does not have the capability to move. 
The framework is primarily evaluated with web-mined clips to demonstrate the effectiveness of \textit{generic} action localization where the primary action can be from both human and non-human actors. A scene is projected onto a screen of dimension $221.4cm \times 124.5cm$, and the camera is positioned such that it can observe only $30\%$ percent of the screen at any given time. A visualization of the evaluation setup, the scene projected, and the scene as observed by the active camera is shown in Figure~\ref{fig:environment}. It can be seen that the camera has to be manipulated effectively to keep track of the action of interest within its field of view. This setup allows us to quantitatively evaluate the ability of our framework to locate and track the most dominant action in a scene under varying conditions of occlusion, motion, and actions.
We also qualitatively evaluate with human and non-human actors that perform generic actions in real-world scenarios. 

\begin{figure*}
    \centering
    \begin{tabular}{ccccc}
         \hline
         \multicolumn{5}{c}{\textit{Sharply Turning Object}}\\\hline
         \includegraphics[width=0.115\textwidth]{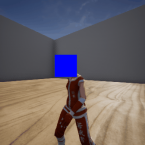} & 
         \includegraphics[width=0.115\textwidth]{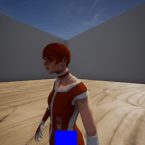} & 
         \includegraphics[width=0.115\textwidth]{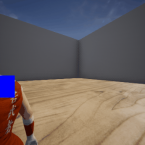} & 
         \includegraphics[width=0.115\textwidth]{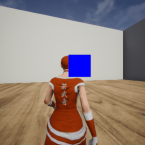} &
         \includegraphics[width=0.115\textwidth]{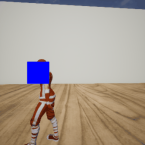} \\
         \hline
         \multicolumn{5}{c}{\textit{Ignoring Clutter with Multiple Objects in Scene}}\\\hline
         \includegraphics[width=0.115\textwidth]{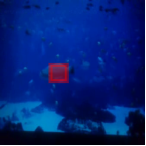} & 
         \includegraphics[width=0.115\textwidth]{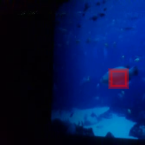} & 
         \includegraphics[width=0.115\textwidth]{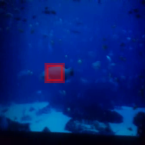} & 
         \includegraphics[width=0.115\textwidth]{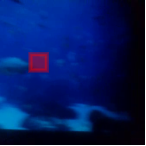} & 
         \includegraphics[width=0.115\textwidth]{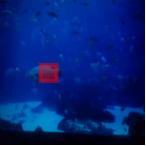} \\
         \hline
         \multicolumn{5}{c}{\textit{Rapidly Changing Environment and Non-Human Actor}}\\\hline
         \includegraphics[width=0.115\textwidth]{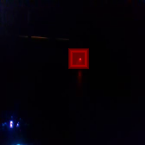} & 
         \includegraphics[width=0.115\textwidth]{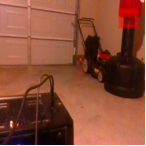} & 
         \includegraphics[width=0.115\textwidth]{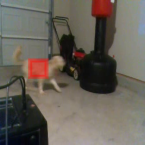} & 
         \includegraphics[width=0.115\textwidth]{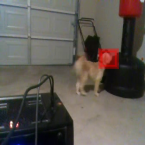} & 
         \includegraphics[width=0.115\textwidth]{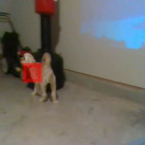} \\
        \hline
         \multicolumn{5}{c}{\textit{Tracking Failure Due to Overshooting Control}}\\\hline
         \includegraphics[width=0.115\textwidth]{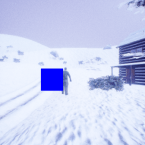} & 
         \includegraphics[width=0.115\textwidth]{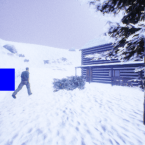} & 
         \includegraphics[width=0.115\textwidth]{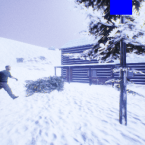} & 
         \includegraphics[width=0.115\textwidth]{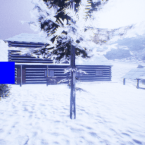} & 
         \includegraphics[width=0.115\textwidth]{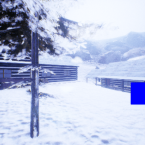} \\
        \hline
         \multicolumn{5}{c}{\textit{Split Attention with Multiple ``Dominant'' Actions}}\\\hline
         \includegraphics[width=0.115\textwidth]{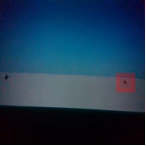} & 
         \includegraphics[width=0.115\textwidth]{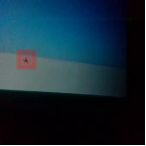} & 
         \includegraphics[width=0.115\textwidth]{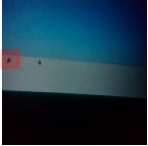} & 
         \includegraphics[width=0.115\textwidth]{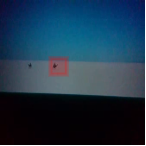} & 
         \includegraphics[width=0.115\textwidth]{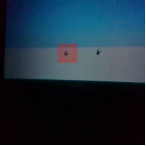} \\
    \end{tabular}
    
    \caption{\textbf{Qualitative illustration} of various success modes (top 3 rows) and failure modes (bottom 2 rows). The attention location of the framework is shown in blue (for simulated environments) and red (for real-world environments).}
    \label{fig:qualitative}
\end{figure*}

\textbf{Data.} To evaluate our framework, we use a collection of $50$ web-mined videos of extreme long-shot sequences in popular movies. Extreme long shot sequences are typically used to establish a location or a scene and are usually in an outdoor location. The major element in the scene would be the background, with some actions occurring in the foreground. Some example scenes that we use in our experiments are shown in Figure~\ref{fig:qualitative}. It can be seen that the environments are busy with multiple actions and sparsely populated actors. 
To annotate actions for evaluation, we ask two users to independently identify up to $k=3$ most dominant actions in the scene by placing a point at the center of the action. In total, there are $127$ distinct actions in our dataset with an average length of $17$ seconds.

\begin{table}[]
    \centering
    \begin{tabular}{|c|c|c|}
    \hline
        \textbf{Approach} & \textbf{AAE} $\downarrow$ & \textbf{AUC} $\uparrow$\\\hline
        Action Oracle & \textbf{3.39} & \textbf{94.38} \\\hline\hline
        Random Prediction & 18.17 & 77.92 \\\hline
        Active TLD Tracker & 14.67 & 79.35 \\\hline
        Active MIL Tracker & 15.72 & 77.68 \\\hline
        Active MOSSE Tracker & 14.81 & 81.34 \\\hline\hline
        Ours & \textbf{9.14} & \textbf{91.12} \\\hline
    \end{tabular}
    \caption{\textbf{Active Action Localization Results.} We report the Average Angular Error (AAE) and average Area Under the Curve (AUC). 
    We outperform all unsupervised approaches by considerable margins while the action oracle provides an upper bound.
    }
    \label{tab:action_localization_perf}
\end{table}

\textbf{Metrics.} It can be hard to quantitatively evaluate generic action localization since there can exist multiple actions in the scene, and the model can pick any one action to actively track and localize. To account for this, we leverage quantitative evaluation metrics from the egocentric gaze prediction task~\cite{itti2006bayesian,fathi2012learning,li2013learning}, which is focused on evaluating how well a computational model can predict the gaze pattern of a human. 
Given that the human gaze is focused on the dominant (or salient) action in a scene, the tasks can be considered to be similar. In this work, we consider the center of the field of view of the active camera as the predicted ``gaze position'' and the actor's position as the ground truth location. If multiple actions are present, we take the closest action as ground truth. 
We use two metrics - \textbf{Average Angular Error} (AAE) and \textbf{Area Under the Curve} (AUC)~\cite{riche2013saliency} as our primary evaluation metrics. The average angular error metric measures the angular distance between the location of the action and the ``gaze position''. 
The other metric, Area Under the Curve (AUC), uses the heatmap of predicted gaze positions as confidence measures to construct an ROC curve. The area under this heat map-based ROC curve is used as the metric to evaluate the saliency. We use the AUC-Judd~\cite{riche2013saliency} version for evaluation. 
The prediction error ($\mathcal{L}_{global}$) is used as a saliency map for evaluation. 

\textbf{Baselines.} We formulate and evaluate a variety of baselines to compare our framework on this difficult benchmark. First, we devise two benchmarks - \textit{Random Prediction} and \textit{Action Oracle}, as our lower and upper bound numbers for our setup. The random prediction model, as the name suggests, takes a random location around the center of the screen and controls the PID controller to center the camera around that location. The action oracle model has access to the ground truth movements of the action and hence positions itself to the correct location. Since the ground truth movement is known, it is constrained only by the hardware to move to the correct location and forms the upper bound. In addition to these baselines, we extend the traditional trackers (described in Section~\ref{sec:aot}) to this task.
We initialize all trackers with region proposals from an EdgeBox detector~\cite{zhu2015tracking} to remove any actor-specific constraints.

\begin{table}[]
    \centering
    \resizebox{\columnwidth}{!}{
    \begin{tabular}{|c|c|c|}
    \hline
    \textbf{Approach} & \textbf{Recall} & \textbf{Precision}\\
    \hline
    PredLearn~\cite{aakur2020action} + PID & 0.69 & 0.16 \\
    \textit{Ours} w/o Normalized Motion & 0.79 & 0.21 \\
    \textit{Ours} w/o $\mathcal{L}_{attn}$ & 0.77 & 0.18 \\
    \textit{Ours} w/ ResNet-50 encoder & 0.86 & 0.27\\
    \hline
    \textit{Ours} w/ $\lambda_p=0.5, \lambda_d=0.1$ & 0.55 & 0.17 \\
    \textit{Ours} w/ $\lambda_p=1, \lambda_d=0$ & 0.80 & 0.22 \\
    \textit{Ours} w/ $\lambda_p=1, \lambda_d=1$ & 0.80 & 0.20 \\
    \hline
    \textit{Full Model} & \textbf{0.85} & \textbf{0.26}\\
    \hline
    \end{tabular}
    }
    \caption{\textbf{Ablation Studies.} We independently evaluate each component's effect on the overall performance.}
    \label{tab:ablation}
\end{table}

\textbf{Quantitative Analysis.} 
We evaluate our framework and the baselines on the real-world robot and summarize results in Table~\ref{tab:action_localization_perf}. It can be seen that the \textit{Action Oracle} model has the best performance with an AAE of $3.39$ and AUC of $94.38\%$. Note that the performance is not perfect, i.e., $0$ and $100\%$, because of hardware constraints. The Braccio arm cannot move as fast as the human eye to keep up with small changes. Although focused on the center of the screen, the random prediction model performs the worst, indicating that just focusing on the center of the screen is not enough in dynamic action scenes. The simulated, pseudo-active trackers (Active TLD, Active MIL, and Active MOSSE) also perform poorly. We note that this performance is similar to what was observed in Section~\ref{sec:aot}, where the trackers would lose track of the object when they make a sharp turn or movement that can place them temporarily out of the field of view of the camera. They fail to recover from such scenarios, although they have access to the exact PID-like mechanism used on our approach. 
We see that our approach performs rather well, obtaining an AAE score of $9.14$ and an AUC score of $91.12$. 
We observe that although the action is not always in the center of the camera's field of view (indicated by a higher AAE), it does place the action within its field of view (as indicated by the AUC score). 
It also exhibits an ability to recover from false movements and distractor actions to an extent (see Section~\ref{sec:qualitative}). 
We notice deviations from ideal when a novel scene is presented or when there are large amounts of clutter. 

\subsection{Ablation Studies}\label{sec:ablation}
We perform ablation studies on the Active Tracking under the \textit{DuelingRoom} environment to assess the contribution of each individual component. We summarize the results in Table~\ref{tab:ablation}. We see that just using predictive learning (as proposed in Aakur \textit{et al.}~\cite{aakur2020action}) with a PID controller does not perform as well on the active vision tasks. The recall and precision are considerably lower. Similarly, using normalized motion (second term of Equation~\ref{eqn:globalPred}) greatly improves the performance. The biggest improvement is provided by the prediction-based feedback for attention (Equation~\ref{eqn:attentionError}). 
The choice of CNN encoder (VGG-16 was used in the full model) to ResNet-50 improves the performance slightly but also adds computational overhead. 
We also vary the different parameters of the reactive PID controller and see that having no anticipatory control ($\lambda_d=0$ in Equation~\ref{eqn:pid}) hurts the performance, whereas placing too much emphasis ($\lambda_d=1$) causes the controller to overshoot the target. Similarly, having less emphasis on reactive control ($\lambda_p=0.5$) causes it to lose track of the object much faster and reduces tracking performance.

\section{Qualitative Analysis}\label{sec:qualitative}
We present some qualitative illustrations in Figure~\ref{fig:qualitative}. In particular, we highlight three significant success modes and two common failure modes of our model. As illustrated in the top row, our embodied agent can keep track of the actor of interest even when they have rapid movements that can place them outside the field of view of the agent. In the second row, we highlight the ability of our model to pick a single \textit{dominant} action in the scene (in this case, the shark swimming through the aquarium) while ignoring other objects in the scene. Similarly, we also evaluate the scenario when the environment rapidly changes, such as when the lighting changes rapidly (as illustrated in the third row). It can be seen from the second column that the attention veers rapidly for a while before it identifies the actor (a non-human one at that) and maintains focus to track and localize it. We also observe two common \textit{failure} cases in our framework. The first, illustrated in the fourth row of Figure~\ref{fig:qualitative}, is that the reactive controller can often overshoot the movement of the active camera and hence lose track of the object of interest since the predictive learning framework can adapt to the new scene and choose a different action to focus on and track. The other common case is the multiple dominant action scenario (shown in the last row of Figure~\ref{fig:qualitative}), where there can exist multiple dominant actions of interest in the scene. Our approach switches attention, focusing on both alternatively based on the actor's predictability. 

\section{Conclusion and Future Work}
In this work, we presented an active vision framework that leverages recent advances in predictive learning for \textit{active} action localization. Using an energy-based formulation, the framework presents promising first steps towards generic action localization in dynamic environments beyond videos. Extensive experiments on simulated and real-world benchmarks show the effectiveness of the proposed approach. We are among the first to demonstrate the effectiveness of active vision for action localization on scenes with complex challenges such as occlusion, noise, camera motion and generic actors, to name a few. We aim to extend our real-world setup into a standard benchmark for fostering research in active vision for event understanding tasks in dynamic environments. 

\textbf{Acknowledgement}
This research was supported in part by the US National Science Foundation grant IIS 1955230.
{\small
\bibliographystyle{ieee_fullname}
\bibliography{egbib}

\begin{thebibliography}{10}\itemsep=-1pt

\bibitem{braccio}
{Tinkerkit Braccio Robot}.
\newblock \url{https://store.arduino.cc/usa/tinkerkit-braccio}.
\newblock Accessed: 2021-06-01.

\bibitem{aakur2020unsupervised}
Sathyanarayanan~N Aakur and Arunkumar Bagavathi.
\newblock Unsupervised gaze prediction in egocentric videos by energy-based
  surprise modeling.
\newblock {\em arXiv preprint arXiv:2001.11580}, 2020.

\bibitem{aakur2019perceptual}
Sathyanarayanan~N Aakur and Sudeep Sarkar.
\newblock A perceptual prediction framework for self supervised event
  segmentation.
\newblock In {\em Proceedings of the IEEE/CVF Conference on Computer Vision and
  Pattern Recognition}, pages 1197--1206, 2019.

\bibitem{aakur2020action}
Sathyanarayanan~N Aakur and Sudeep Sarkar.
\newblock Action localization through continual predictive learning.
\newblock {\em arXiv preprint arXiv:2003.12185}, 2020.

\bibitem{aakur2021learning}
Sathyanarayanan~N Aakur and Sudeep Sarkar.
\newblock Learning actor-centered representations for action localization in
  streaming videos using predictive learning.
\newblock {\em arXiv preprint arXiv:2104.14131}, 2021.

\bibitem{alzarok20173d}
Hamza Alzarok, Simon Fletcher, and Andrew~P Longstaff.
\newblock 3d visual tracking of an articulated robot in precision automated
  tasks.
\newblock {\em Sensors}, 17(1):104, 2017.

\bibitem{araki2009pid}
M Araki.
\newblock Pid control.
\newblock {\em Control Systems, Robotics and Automation: System Analysis and
  Control: Classical Approaches II}, pages 58--79, 2009.

\bibitem{babenko2009visual}
Boris Babenko, Ming-Hsuan Yang, and Serge Belongie.
\newblock Visual tracking with online multiple instance learning.
\newblock In {\em 2009 IEEE Conference on Computer Vision and Pattern
  Recognition}, pages 983--990. IEEE, 2009.

\bibitem{bahdanau2016neural}
Dzmitry Bahdanau, Kyunghyun Cho, and Yoshua Bengio.
\newblock Neural machine translation by jointly learning to align and
  translate, 2016.

\bibitem{bolme2010visual}
David~S Bolme, J~Ross Beveridge, Bruce~A Draper, and Yui~Man Lui.
\newblock Visual object tracking using adaptive correlation filters.
\newblock In {\em 2010 IEEE Conference on Computer Vision and Pattern
  Recognition}, pages 2544--2550. IEEE, 2010.

\bibitem{brockman2016openai}
Greg Brockman, Vicki Cheung, Ludwig Pettersson, Jonas Schneider, John Schulman,
  Jie Tang, and Wojciech Zaremba.
\newblock Openai gym.
\newblock {\em arXiv preprint arXiv:1606.01540}, 2016.

\bibitem{brooks1986robust}
Rodney Brooks.
\newblock A robust layered control system for a mobile robot.
\newblock {\em IEEE Journal on Robotics and Automation}, 2(1):14--23, 1986.

\bibitem{ccelik2017color}
Yunus {\c{C}}elik, Mahmut Altun, and Mahit G{\"u}ne{\c{s}}.
\newblock Color based moving object tracking with an active camera using motion
  information.
\newblock In {\em 2017 International Artificial Intelligence and Data
  Processing Symposium (IDAP)}, pages 1--4. IEEE, 2017.

\bibitem{crowley1994integration}
James~L Crowley, Jean~Marc Bedrune, Morten Bekker, and Michael Schneider.
\newblock Integration and control of reactive visual processes.
\newblock In {\em European Conference on Computer Vision}, pages 47--58.
  Springer, 1994.

\bibitem{daniilidis1998real}
Kostas Daniilidis, Christian Krauss, Michael Hansen, and Gerald Sommer.
\newblock Real-time tracking of moving objects with an active camera.
\newblock {\em Real-Time Imaging}, 4(1):3--20, 1998.

\bibitem{DBLP:journals/corr/abs-1711-11543}
Abhishek Das, Samyak Datta, Georgia Gkioxari, Stefan Lee, Devi Parikh, and
  Dhruv Batra.
\newblock Embodied question answering.
\newblock {\em CoRR}, abs/1711.11543, 2017.

\bibitem{5206848}
Jia Deng, Wei Dong, Richard Socher, Li-Jia Li, Kai Li, and Li Fei-Fei.
\newblock Imagenet: A large-scale hierarchical image database.
\newblock In {\em 2009 IEEE Conference on Computer Vision and Pattern
  Recognition}, pages 248--255, 2009.

\bibitem{dickinson1997active}
Sven~J Dickinson, Henrik~I Christensen, John~K Tsotsos, and G{\"o}ran Olofsson.
\newblock Active object recognition integrating attention and viewpoint
  control.
\newblock {\em Computer Vision and Image Understanding}, 67(3):239--260, 1997.

\bibitem{dickinson1993integrating}
Sven~J Dickinson, Suzanne Stevenson, Eugene Amdur, John~K Tsotsos, and Lars
  Olsson.
\newblock Integrating task-directed planning with reactive object recognition.
\newblock In {\em Intelligent Robots and Computer Vision XII: Algorithms and
  Techniques}, volume 2055, pages 212--224. International Society for Optics
  and Photonics, 1993.

\bibitem{duarte2018videocapsulenet}
Kevin Duarte, Yogesh~S Rawat, and Mubarak Shah.
\newblock Videocapsulenet: A simplified network for action detection.
\newblock {\em arXiv preprint arXiv:1805.08162}, 2018.

\bibitem{fathi2012learning}
Alireza Fathi, Yin Li, and James~M Rehg.
\newblock Learning to recognize daily actions using gaze.
\newblock In {\em European Conference on Computer Vision}, pages 314--327.
  Springer, 2012.

\bibitem{gkioxari2015finding}
Georgia Gkioxari and Jitendra Malik.
\newblock Finding action tubes.
\newblock In {\em Proceedings of the IEEE Conference on Computer Vision and
  Pattern Recognition}, pages 759--768, 2015.

\bibitem{grunnet2018projectors}
Anders Grunnet-Jepsen, John~N Sweetser, Paul Winer, Akihiro Takagi, and John
  Woodfill.
\newblock Projectors for intel{\textregistered} realsense™ depth cameras
  d4xx.
\newblock {\em Intel Support, Interl Corporation: Santa Clara, CA, USA}, 2018.

\bibitem{hochreiter1997long}
Sepp Hochreiter and J{\"u}rgen Schmidhuber.
\newblock Long short-term memory.
\newblock {\em Neural Computation}, 9(8):1735--1780, 1997.

\bibitem{horstmann2015surprise}
Gernot Horstmann and Arvid Herwig.
\newblock Surprise attracts the eyes and binds the gaze.
\newblock {\em Psychonomic Bulletin \& Review}, 22(3):743--749, 2015.

\bibitem{horstmann2016novelty}
Gernot Horstmann and Arvid Herwig.
\newblock Novelty biases attention and gaze in a surprise trial.
\newblock {\em Attention, Perception, \& Psychophysics}, 78(1):69--77, 2016.

\bibitem{hou2017tube}
Rui Hou, Chen Chen, and Mubarak Shah.
\newblock Tube convolutional neural network (t-cnn) for action detection in
  videos.
\newblock In {\em Proceedings of the IEEE International Conference on Computer
  Vision (ICCV)}, pages 5822--5831, 2017.

\bibitem{huang2019learning}
Chong Huang, Zhenyu Yang, Yan Kong, Peng Chen, Xin Yang, and Kwang-Ting~Tim
  Cheng.
\newblock Learning to capture a film-look video with a camera drone.
\newblock In {\em 2019 International Conference on Robotics and Automation
  (ICRA)}, pages 1871--1877. IEEE, 2019.

\bibitem{itti2006bayesian}
Laurent Itti and Pierre~F Baldi.
\newblock Bayesian surprise attracts human attention.
\newblock In {\em Advances in Neural Information Processing Systems}, pages
  547--554, 2006.

\bibitem{jain2017tubelets}
Mihir Jain, Jan Van~Gemert, Herv{\'e} J{\'e}gou, Patrick Bouthemy, and Cees~GM
  Snoek.
\newblock Tubelets: Unsupervised action proposals from spatiotemporal
  super-voxels.
\newblock {\em International Journal of Computer Vision}, 124(3):287--311,
  2017.

\bibitem{kalal2011tracking}
Zdenek Kalal, Krystian Mikolajczyk, and Jiri Matas.
\newblock Tracking-learning-detection.
\newblock {\em IEEE Transactions on Pattern Analysis and Machine Intelligence},
  34(7):1409--1422, 2011.

\bibitem{kolve2019ai2thor}
Eric Kolve, Roozbeh Mottaghi, Winson Han, Eli VanderBilt, Luca Weihs, Alvaro
  Herrasti, Daniel Gordon, Yuke Zhu, Abhinav Gupta, and Ali Farhadi.
\newblock Ai2-thor: An interactive 3d environment for visual ai, 2019.

\bibitem{kyrkou2020imitation}
Christos Kyrkou.
\newblock Imitation-based active camera control with deep convolutional neural
  network.
\newblock {\em arXiv preprint arXiv:2012.06428}, 2020.

\bibitem{li2020pose}
Jing Li, Jing Xu, Fangwei Zhong, Xiangyu Kong, Yu Qiao, and Yizhou Wang.
\newblock Pose-assisted multi-camera collaboration for active object tracking.
\newblock In {\em Proceedings of the AAAI Conference on Artificial
  Intelligence}, volume~34, pages 759--766, 2020.

\bibitem{li2013learning}
Yin Li, Alireza Fathi, and James~M Rehg.
\newblock Learning to predict gaze in egocentric video.
\newblock In {\em Proceedings of the IEEE International Conference on Computer
  Vision}, pages 3216--3223, 2013.

\bibitem{DBLP:journals/corr/LuoSML17}
Wenhan Luo, Peng Sun, Yadong Mu, and Wei Liu.
\newblock End-to-end active object tracking via reinforcement learning.
\newblock {\em CoRR}, abs/1705.10561, 2017.

\bibitem{qiu2017unrealcv}
Weichao Qiu, Fangwei Zhong, Yi Zhang, Siyuan Qiao, Zihao Xiao, Tae~Soo Kim, and
  Yizhou Wang.
\newblock Unrealcv: Virtual worlds for computer vision.
\newblock In {\em Proceedings of the 25th ACM International Conference on
  Multimedia}, pages 1221--1224, 2017.

\bibitem{sdk2021}
Santhosh~K. Ramakrishnan, Dinesh Jayaraman, and Kristen Grauman.
\newblock An exploration of embodied visual exploration.
\newblock {\em International Journal of Computer Vision}, 129:1616--1649, 2021.

\bibitem{ri2017image}
Yoshi Ri and Hiroshi Fujimoto.
\newblock Image based visual servo application on video tracking with monocular
  camera based on phase correlation method.
\newblock In {\em IEEJ International Workshop on Sensing, Actuation, Motion
  Control, and Optimization}, 2017.

\bibitem{riche2013saliency}
Nicolas Riche, Matthieu Duvinage, Matei Mancas, Bernard Gosselin, and Thierry
  Dutoit.
\newblock Saliency and human fixations: State-of-the-art and study of
  comparison metrics.
\newblock In {\em Proceedings of the IEEE International Conference on Computer
  Vision}, pages 1153--1160, 2013.

\bibitem{habitat19iccv}
Manolis Savva, Abhishek Kadian, Oleksandr Maksymets, Yili Zhao, Erik Wijmans,
  Bhavana Jain, Julian Straub, Jia Liu, Vladlen Koltun, Jitendra Malik, Devi
  Parikh, and Dhruv Batra.
\newblock Habitat: {A} {P}latform for {E}mbodied {AI} {R}esearch.
\newblock In {\em Proceedings of the IEEE/CVF International Conference on
  Computer Vision (ICCV)}, 2019.

\bibitem{shenigibson}
Bokui Shen, Fei Xia, Chengshu Li, Roberto Mart{\i}n-Mart{\i}n, Linxi Fan,
  Guanzhi Wang, Shyamal Buch, Claudia D’Arpino, Sanjana Srivastava, Lyne~P
  Tchapmi, Kent Vainio, Li Fei-Fei, and Silvio Savarese.
\newblock igibson, a simulation environment for interactive tasks in large
  realistic scenes.
\newblock {\em arXiv preprint}, 2020.

\bibitem{simonyan2015deep}
Karen Simonyan and Andrew Zisserman.
\newblock Very deep convolutional networks for large-scale image recognition,
  2015.

\bibitem{soomro2015action}
Khurram Soomro, Haroon Idrees, and Mubarak Shah.
\newblock Action localization in videos through context walk.
\newblock In {\em Proceedings of the IEEE International Conference on Computer
  Vision}, pages 3280--3288, 2015.

\bibitem{soomro2016predicting}
Khurram Soomro, Haroon Idrees, and Mubarak Shah.
\newblock Predicting the where and what of actors and actions through online
  action localization.
\newblock In {\em Proceedings of the IEEE Conference on Computer Vision and
  Pattern Recognition}, pages 2648--2657, 2016.

\bibitem{spurlock2015discriminative}
Scott Spurlock, Junjie Shan, and Richard Souvenir.
\newblock Discriminative poses for early recognition in multi-camera networks.
\newblock In {\em Proceedings of the 9th International Conference on
  Distributed Smart Cameras}, pages 74--79, 2015.

\bibitem{wang2020active}
Boyu Wang, Lihan Huang, and Minh Hoai.
\newblock Active vision for early recognition of human actions.
\newblock In {\em Proceedings of the IEEE/CVF Conference on Computer Vision and
  Pattern Recognition}, pages 1081--1091, 2020.

\bibitem{wang2018action}
Pichao Wang, Wanqing Li, Chuankun Li, and Yonghong Hou.
\newblock Action recognition based on joint trajectory maps with convolutional
  neural networks.
\newblock {\em Knowledge-Based Systems}, 158:43--53, 2018.

\bibitem{wilkes1993behaviors}
David~R Wilkes and John~K Tsotsos.
\newblock Behaviors for active object recognition.
\newblock In {\em Intelligent Robots and Computer Vision XII: Algorithms and
  Techniques}, volume 2055, pages 225--239. International Society for Optics
  and Photonics, 1993.

\bibitem{xie2018correlation}
Yanchun Xie, Jimin Xiao, Kaizhu Huang, Jeyarajan Thiyagalingam, and Yao Zhao.
\newblock Correlation filter selection for visual tracking using reinforcement
  learning.
\newblock {\em IEEE Transactions on Circuits and Systems for Video Technology},
  30(1):192--204, 2018.

\bibitem{zacks2001event}
Jeffrey~M Zacks and Barbara Tversky.
\newblock Event structure in perception and conception.
\newblock {\em Psychological Bulletin}, 127(1):3, 2001.

\bibitem{zhong2017gym}
Fangwei Zhong, Weichao Qiu, Tingyun Yan, Y Alan, and Yizhou Wang.
\newblock Gym-unrealcv: Realistic virtual worlds for visual reinforcement
  learning.
\newblock {\em Web Page}, 2017.

\bibitem{zhong2018ad}
Fangwei Zhong, Peng Sun, Wenhan Luo, Tingyun Yan, and Yizhou Wang.
\newblock Ad-vat: An asymmetric dueling mechanism for learning visual active
  tracking.
\newblock In {\em International Conference on Learning Representations}, 2018.

\bibitem{zhu2015tracking}
Gao Zhu, Fatih Porikli, and Hongdong Li.
\newblock Tracking randomly moving objects on edge box proposals.
\newblock {\em arXiv preprint arXiv:1507.08085}, 2015.

\end{thebibliography}
}

\end{document}